\documentclass[fleqn,10pt]{wlscirep}
\usepackage[utf8]{inputenc}
\usepackage[T1]{fontenc}
\usepackage{soul}
\usepackage{multirow}
\usepackage{CJKutf8}

\usepackage{graphicx}
\usepackage{tabularx}
\usepackage{subfigure}
\usepackage{subcaption}
\usepackage{amsmath}
\usepackage{makecell}
\usepackage{multirow}
\usepackage{arydshln}
\usepackage{algorithm}
\usepackage{algorithmic}
\usepackage{float}
\usepackage[normalem]{ulem}
\usepackage{array}
\newcolumntype{C}[1]{>{\centering\arraybackslash}m{#1}}
\useunder{\uline}{\ul}{}
\usepackage{listings}
\title{GRASP: A Grid-Based Benchmark for Evaluating Commonsense Spatial Reasoning}

\author[1]{Zhisheng Tang}
\author[1]{Mayank Kejriwal}

\affil[1]{University of Southern California, Information Sciences Institute, Marina del Rey, 90292, United States of America}

\affil[*]{kejriwal@isi.edu}

\begin{abstract}

Spatial reasoning, an important faculty of human cognition with many practical applications, is one of the core commonsense skills that is not purely language-based and, for satisfying (as opposed to optimal) solutions, requires some minimum degree of planning. Existing benchmarks of Commonsense Spatial Reasoning (CSR) tend to evaluate how Large Language Models (LLMs) interpret text-based spatial \textit{descriptions} rather than directly evaluate a plan produced by the LLM in response to a \textit{specific} spatial reasoning problem. In this paper, we construct a large-scale benchmark called GRASP, which consists of 16,000 grid-based environments where the agent is tasked with an energy collection problem. These environments include 100 grid instances instantiated using each of the 160 different grid settings, involving five different energy distributions, two modes of agent starting position, and two distinct obstacle configurations, as well as three kinds of agent constraints. Using GRASP, we compare classic baseline approaches, such as random walk and greedy search methods, with advanced LLMs like GPT-3.5-Turbo, GPT-4o, and GPT-o1-mini. The experimental results indicate that even these advanced LLMs struggle to consistently achieve satisfactory solutions.

\end{abstract}
\begin{document}

\flushbottom
\maketitle

\thispagestyle{empty}

% \section*{Introduction}

Spatial reasoning, a fundamental aspect of human cognition, is a core commonsense skill that allows people to navigate, interact with, and manipulate their environments effectively \cite{davis2011theory, newcombe2010early, ishikawa2021spatial, hegarty2005individual}. This cognitive ability involves recognizing and understanding the spatial relations among different entities, \cite{pellegrino1984understanding}, as well as actively using it to navigate through space while manipulating objects within it \cite{lohman1979spatial}, which is essential in day-to-day activities such as walking and driving. An adequate spatial reasoning ability means more than just recognizing the relative spatial relationships between objects but also using such relationships, in combination with planning skills, to achieve satisfying, if not optimal, solutions for tasks at hand.

With the advancement of Large Language Models (LLMs) in understanding and generating natural language \cite{achiam2023gpt, zhao2023survey, wei2022emergent, touvron2023llama}, exhibiting near human-level commonsense reasoning abilities \cite{li2021systematic, zhao2024large, richardson2023commonsense}, as well as their increasing integration into applications requiring spatial awareness \cite{gallotta2024large, ccelen2024design, yan2023survey, zhou2024large}, evaluating their spatial reasoning capabilities has raised significant attention \cite{sharma2023exploring, wu2024visualization, yamada2023evaluating, aghzal2023can}. Several benchmarks, such as SpartQA \cite{mirzaee2021spartqa}, StepGame\cite{shi2022stepgame}, and BabyAI\cite{chevalier2018babyai}, have been proposed to evaluate these abilities by simulating visual or textual spatial relationships. However, these benchmarks focus on interpreting spatial descriptions, whether textually or visually, and neglect the practical aspect of spatial reasoning. 

The ability to use interpreted spatial information to navigate and interact with the environment in a commonsensical manner is termed Commonsense Spatial Reasoning (CSR). It is essential to effectively navigate around environments. For example, an agent must understand the fundamental spatial principles such as `if an obstacle is in the way, an alternative path that goes around the obstacle should be considered' and `if two resources are close together, it is more efficient to collect them in one trip rather than multiple trips.' Such CSR enables the agent to make effective decisions with regard to the spatial layout, such as prioritizing the collection of nearby resources before exploring further and adapting to obstacles by rerouting effectively. 

We introduce a novel benchmark, GRASP, that evaluates the CSR abilities of LLMs within a structured grid environment. Unlike previous spatial commonsense datasets and benchmarks \cite{liu2022things, lin2023spatial, zhang2020language, cui2020beyond}, which focus on evaluating the ability to understand the relative scales and size of objects, GRASP emphasizes practical applications of spatial reasoning, investigating the ability to use and reason through spatial information in a commonsensical manner. GRASP consists of grid-based environments with an agent in it. The task of the agent in this environment is to navigate through the grid, avoiding potential obstacles, collect resources (i.e., energy) along the way and get back to the starting point to drop the resources. Unlike previous benchmarks that rely heavily on textual descriptions or visual elements to convey spatial information, GRASP directly integrates text renderings of grid environments, which allows for a more direct assessment without any intermediate interpretation steps. The spatial relationships are expressed horizontally using the relative position of the text itself and vertically using strings such as the newline character (i.e., `\textbackslash n'). By focusing on the LLMs' own ability to interpret and act upon spatial information, GRASP assesses LLMs' CSR ability to come up with plans to navigate in the grid while collecting as much energy as possible to drop to the starting point within a fixed number of steps.

GRASP is a set of environments that consists of 2,000 different grid instances and eight distinct combinations of agent constraints, resulting in 16,000 environments. Each grid is constructed as a two-dimensional array of cells, each of which can be empty, has an obstacle, or contains a unit of energy. The objective for the agent is to collect as much energy as possible within a fixed number of steps, simulating real-world scenarios where agents must navigate through the environments and plan a path to collect resources. The environment includes various patterns of energy arrangement, such as random, vertically-skewed, horizontally-skewed, clustered, and spiral distributions. These patterns challenge the agent's ability to recognize the differences and adapt its strategy based on different distributions of resources. Additionally, obstacles are randomly placed within the grid, and the agent's starting position is randomized within specified inner or outer regions of the grid, testing its adaptability to different starting scenarios. Finally, we introduce various constraints that affect the behaviors of the agent, starting with a limited set of actions that include basic movements and, in a more complex setup, additional diagonal movements. Additional constraints, such as the maximum number of energy units the agent can carry and the cost associated with each step, are designed to further test the agent's CSR and associated planning abilities.

We use two baseline agents to set a foundation for evaluating the performance of LLM agents in the GRASP benchmark. Experimental results reveal that GPT-3.5-Turbo, GPT-4o, and GPT-o1-mini all underperform compared to the first baseline method, which is based on greedy search. We also find that GPT-3.5-Turbo  falls short of the second, simpler random walk baseline. Among the LLM-based agents, GPT-o1-mini exhibits the most promising performance, achieving, on average, the best results and demonstrating reasonable adaptability to constraints such as obstacles and limited allowed movements. However, the introduction of constraints such as energy costs per step or energy-carrying limit significantly impacts all LLM-based agents' performance, showing that there are still significant challenges to deploying them on reasoning and planning problems that resemble CSR. Our findings ultimately support GRASP’s potential to benchmark and reveal important differences in agents’ CSR abilities and offer valuable insights into the capability and shortcomings of LLM-based agents in practical spatial reasoning tasks.

% \section*{Results}\label{results}

\section*{Results}

\subsection*{Benchmark Overview}
\begin{figure}[h]
\includegraphics[width=\textwidth]{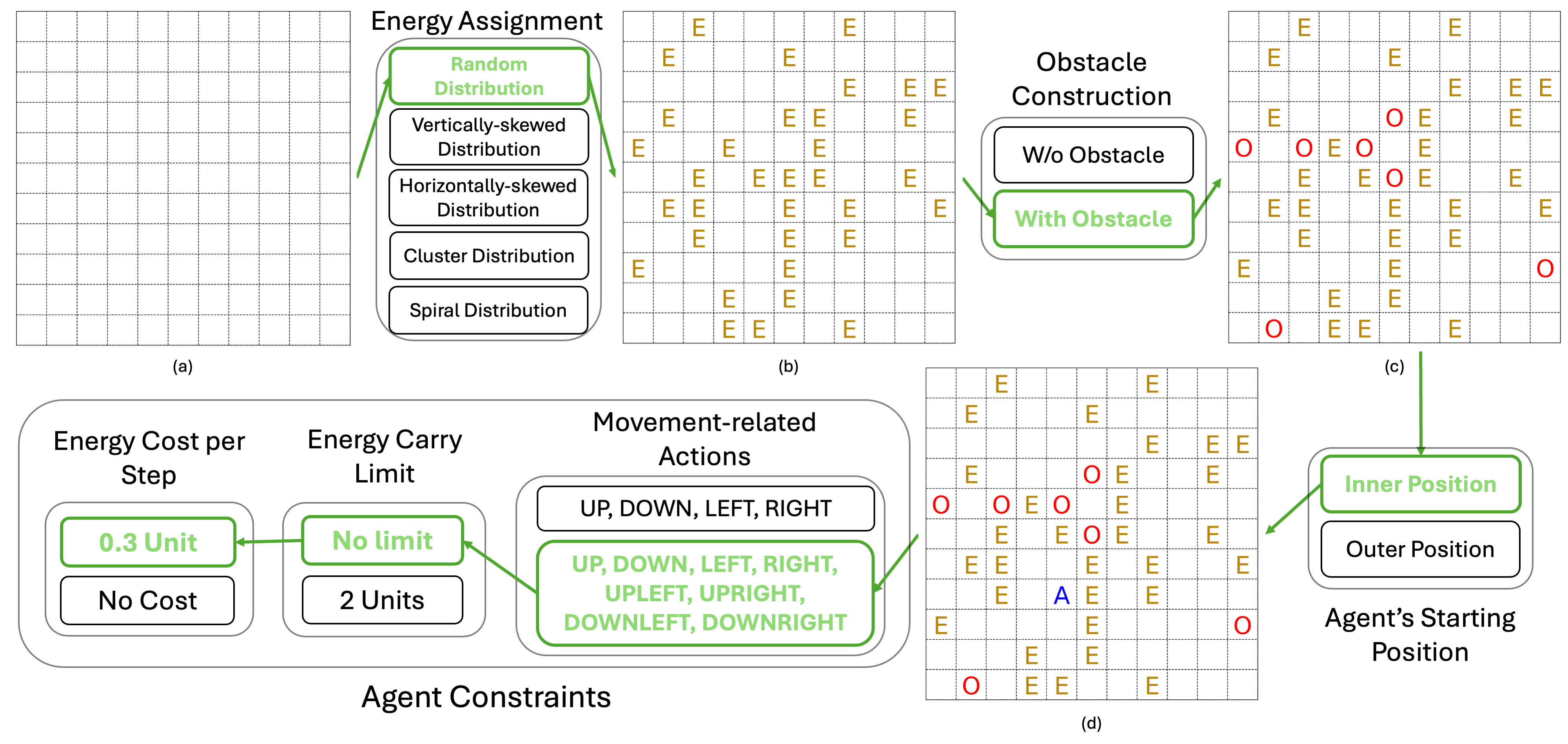}
\caption{
The pipeline for constructing one instance of the grid environment in the GRASP benchmark.
}\label{fig:bench_con}
\end{figure}

We designed GRASP specifically to benchmark LLM-based agents' CSR abilities. GRASP tests LLM-based agents' capability to navigate in a grid-based environment, represented using simple text characters, while assigned an energy collection task. The goal of the agent is to collect as much energy as possible within a certain number of steps. Figure \ref{fig:bench_con} shows the entire pipeline for constructing one instance of the grid environment. The benchmark is a combination of grid environments and agent constraints. For one grid environment, there can be five different types of energy distribution: \textit{random, vertically-skewed, horizontally-skewed, cluster,} and \textit{spiral}. These five types are used to create a broad enough set of scenarios that cannot be `gamed' by agents through memorization or obvious heuristics. Grid environments can also contain obstacles that block specific grid cells, making those cells inaccessible. The obstacles are randomly placed inside the grids to introduce challenges and encourage strategic navigation. The agent's starting position can be either the inner position or the outer position. The inner position refers to cells near the center of the grid, while the outer position is the cells near the edges. These starting positions allow the benchmark to evaluate the agent’s performance under different spatial conditions.

There are three kinds of agent constraints:
\begin{enumerate}
    \item First, the agent is allowed two types of \textit{movement-related actions}: one with four directions and the other with eight directions. These movement options assess the agent’s ability to navigate the grid with varying levels of freedom.
    \item Second, the agent can have an \textit{energy carry limit}, which restricts the amount of energy the agent can hold at any given time. This constraint encourages the agent to plan its energy collection and usage strategically.
    \item Finally, the agent can incur an \textit{energy cost per step}, which determines the amount of energy consumed with each step taken. This cost introduces a trade-off between exploration and resource conservation, challenging the agent to optimize its movements. The agent is also restricted to only 20 steps. The detailed parameters and processes are described in \textit{Methods}.
\end{enumerate}  

\subsection*{Experimental Setup and Parameters}
We start by briefly describing the parameters of the experimental study for evaluating CSR in LLM-based agents using GRASP. 

\textbf{Baselines:} To evaluate agent performance on GRASP, we begin by establishing two baseline methods that represent different levels of capabilities. The baselines establish reference points for interpreting the results of the LLM-based agents. The first baseline is the \textit{random walk} method. It represents a lower-bound performance and serves as a control to assess the inherent difficulty of the task. It simulates an agent who has no knowledge about the grid, the agent constraints regarding the limit on the number of energy that the agent can carry, and the cost of each action, but is only aware of the set of available actions. An agent who employs this approach randomly selects actions from the set of available movements. Basically, this baseline represents how a completely uninformed agent would perform in GRASP, offering a minimum performance level against which more sophisticated agents can be compared. 

The second baseline is the \textit{greedy search} method, which continues to be popular in many practical planning scenarios\cite{greedy1,greedy2,greedy3} where devising optimal solutions (e.g., by solving an elaborately specified integer linear program) is combinatorially explosive. This method simulates an agent with perfect knowledge of the grid environment but no planning abilities and no knowledge about the agent constraints except the available movement-related actions. The greedy agent always selects the action that maximizes immediate energy gain. This baseline demonstrates how full environmental knowledge can enhance performance. Implementation details are provided in \textit{Methods}.

\textbf{LLM-based agents:} In addition to the baseline approaches, we evaluate the performance of agents based on three LLMs: GPT-3.5-Turbo (i.e., gpt-3.5-turbo-0125), GPT-4o, and GPT-o1-mini. All three models are commercial, with GPT-4o and GPT-o1-mini being close to the state-of-the-art at the time of writing. The models were chosen due to their varying levels of capability and costs. GPT-3.5-Turbo is included as a cost-effective, widely accessible model, while GPT-4o is evaluated as a more advanced model known for its superior reasoning capabilities. GPT-o1-mini is both more expensive and computationally stronger compared to GPT-3.5-Turbo and GPT-4o, making it an interesting point of comparison. The first two models are used with a temperature setting of 0 to ensure reproducibility. However, because GPT-o1-mini does not allow modification to its default temperature, it is kept at 1. 

These models are tested using zero-shot prompting . To construct the prompt used for the LLMs, we utilize all available information, both from the grid environment and agent constraints. Instructions and agent constraints are provided to LLMs in the system prompt, and the grid is provided in the user prompt; the LLMs are required to provide a full list of actions in the response, with no further interaction with the environment. GPT-o1-mini can only accept user prompts, so all information is provided in the user prompt. The specific prompts used are detailed in \textit{Methods}. The experiments are conducted using one-tenth of the entire GRASP benchmark, for grid instances indexed from 0 to 9, except for GPT-o1-mini, which is evaluated using grid instances indexed from 0 to 4.

\subsection*{Performance Evaluation}

\begin{table}[h]\caption{
The average number of steps taken by the agents (displayed in the \textit{Length} column) and the average units of energy collected and dropped to the starting position by the agents (displayed in the \textit{Energy} column), are calculated for each control setting. The green numbers in each row represent the maximum energy collected by every agent, while the red numbers represent the minimum ones.
}\label{tab:results}
\centering
\resizebox{\columnwidth}{!}{
\begin{tabular}{cccccccccccc}
\hline
\multirow{2}{*}{\makecell{$\downarrow$ \textbf{Control} $\&$ \\ \textbf{its Values} $\searrow$}} & \multirow{2}{*}{\textbf{Agent} $\rightarrow$}  &\multicolumn{2}{c}{\makecell{Random \\ Walk}} &\multicolumn{2}{c}{\makecell{Greedy \\ Search}} & \multicolumn{2}{c}{\makecell{GPT-3.5-\\Turbo}} &\multicolumn{2}{c}{GPT-4o} &\multicolumn{2}{c}{GPT-o1-mini} \\
\cline{3-12}
& & \textit{Length} & \textit{Energy} & \textit{Length} & \textit{Energy}& \textit{Length} & \textit{Energy}& \textit{Length} & \textit{Energy} & \textit{Length} & \textit{Energy} \\
\hline
\multirow{5}{*}{\makecell{Energy \\Distribution}} &Random &19.0 &-0.80 &18.8 &\textcolor{green}{0.37} &19.7 &\textcolor{red}{-1.96} &16.6 &-0.99 &18.5 &-0.44 \\
% & 19.0 &-0.97 &18.6 &\textcolor{green}{0.47} &19.7 &\textcolor{red}{-2.11} &17.3 &-1.26 &18.4 &-0.32
% &19.00 &-1.26 &18.75 & \textcolor{green}{-0.14} &19.70 &\textcolor{red}{-1.96} &16.60 &-1.20 \\
\cline{2-2}
&Vertically-skewed &19.0 &-0.98 &18.7 &\textcolor{green}{0.33} &19.6 &\textcolor{red}{-1.66} &17.4 &-1.40 &18.6 &-0.30 \\
% & 19.0 &-1.60 &18.8 &\textcolor{green}{0.37} &18.6 &\textcolor{red}{-2.07} &16.7 &-1.06 &18.2 &-0.33
% &19.00 &-1.43 &18.65 &\textcolor{green}{-0.23} &19.59 &\textcolor{red}{-1.66} &17.40 &-1.55 \\
\cline{2-2}
&Horizontally-skewed & 19.0 &-0.97 &18.6 &\textcolor{green}{0.47} &19.7 &\textcolor{red}{-2.11} &17.3 &-1.26 &18.4 &-0.32 \\
% &19.00 &-1.33 &18.62 &\textcolor{green}{-0.07} &19.74 &\textcolor{red}{-2.12} &17.28 &-1.43\\
\cline{2-2}
&Cluster & 19.0 &-1.60 &18.8 &\textcolor{green}{0.37} &18.6 &\textcolor{red}{-2.07} &16.7 &-1.06 &18.2 &-0.33 \\
% &19.00 &-1.90 &18.81 &\textcolor{green}{-0.09} &18.58 &\textcolor{red}{-2.09} &16.65 &-1.28 \\
\cline{2-2}
&Spiral &19.0 &-1.38 &18.7 &\textcolor{green}{0.20} &19.4 &\textcolor{red}{-2.32} &17.1 &-1.34 &18.3 &-0.17 \\
% &19.00 &-1.77 &18.73 &\textcolor{green}{-0.15} &19.41 &\textcolor{red}{-2.34} &17.09 &-1.45 \\
\hline
\multirow{2}{*}{Obstacle} &Yes &19.0 &-1.18 &18.7 &\textcolor{green}{0.31} &19.3 &\textcolor{red}{-2.08} &17.2 &-1.45 &18.3 &-0.32 \\
% &19.0 &-1.11 &18.7 &\textcolor{green}{0.39} &19.5 &\textcolor{red}{-1.97} &16.8 &-0.96 &18.6 &-0.31
% &19.00 &-1.57 &18.70 &\textcolor{green}{-0.19} &19.33 &\textcolor{red}{-2.09} &17.18 &-1.58 \\
\cline{2-2}
&No &19.0 &-1.11 &18.7 &\textcolor{green}{0.39} &19.5 &\textcolor{red}{-1.97} &16.8 &-0.96 &18.6 &-0.31 \\
% &19.00 &-1.51 &18.72 &\textcolor{green}{-0.09} &19.48 &\textcolor{red}{-1.98} &16.83 &-1.18 \\
\hline
\multirow{2}{*}{\makecell{Starting Position}} &Inner Position &19.0 &-1.06 &18.7 &\textcolor{green}{0.46} &19.4 &\textcolor{red}{-1.90} &17.0 &-1.17 &18.7 &-0.35 \\
% &19.00 &-1.50 &18.73 &\textcolor{green}{-0.02} &19.38 &\textcolor{red}{-1.92} &17.04 &-1.34 \\
\cline{2-2}
&Outer Position &19.0 &-1.23 &18.7 &\textcolor{green}{0.23} &19.4 &\textcolor{red}{-2.14} &17.0 &-1.25 &18.3 &-0.27 \\
% &19.00 &-1.58 &18.70 &\textcolor{green}{-0.25} &19.43 &\textcolor{red}{-2.14} &16.96 &-1.42 \\
\hline
\multirow{2}{*}{\makecell{Movement-related \\ Action Set}} &$\mu_1$ (4 directions)&19.0 &-1.21 &18.5 &\textcolor{green}{0.80} &19.1 &\textcolor{red}{-1.89} &16.8 &-1.13 &18.5 &-0.19 \\
% &19.00 &-1.21 &18.54 &\textcolor{green}{0.80} &19.12 &\textcolor{red}{-1.89} &16.75 &-1.13 \\
\cline{2-2}
&$\mu_2$ (8 directions) &19.0 &-1.08 &18.9 &\textcolor{green}{-0.10} &19.7 &\textcolor{red}{-2.15} &17.3 &-1.28 &18.5 &-0.43 \\
% &19.00 &-1.10 &18.89 &\textcolor{green}{-0.10} &19.69 &\textcolor{red}{-2.14} &17.26 &-1.28 \\
\hline
\multirow{2}{*}{\makecell{Energy \\Carrying Limit}} &No Limit &19.0 &-0.89 &18.7 &\textcolor{green}{1.50} &19.8 &\textcolor{red}{-2.00} &16.7 &-0.85 &18.3 &0.24 \\ 
% &19.00 &-1.40 &18.73 &\textcolor{green}{0.83} &19.78 &\textcolor{red}{-2.00} &16.72 &-1.08 \\
\cline{2-2}
&2 Units &19.0 &-1.40 &18.7 &\textcolor{green}{-0.81} &19.0 &\textcolor{red}{-2.05} &17.3 &-1.57 &18.6 &-0.87 \\
% &19.00 &-1.68 &18.69 &\textcolor{green}{-1.10} &19.03 &\textcolor{red}{-2.06} &17.28 &-1.68 \\
\hline
\multirow{2}{*}{\makecell{Energy Cost \\ Per Step}} &0 Unit &19.0 &1.68 &18.7 &\textcolor{green}{3.14} &19.4 &\textcolor{red}{1.25} &17.2 &1.34 &18.4 &2.39 \\ 
% &19.00 &1.30 &18.72 &\textcolor{green}{2.66} &19.40 &1.24 &17.15 &\textcolor{red}{1.16} \\
\cline{2-2}
&0.3 Unit &19.0 &-3.97 &18.7 &\textcolor{green}{-2.44} &19.4 &\textcolor{red}{-5.30} &16.9 &-3.76 &18.6 &-3.02 \\
% &19.00 &-4.38 &18.70 &\textcolor{green}{-2.93} &19.41 &\textcolor{red}{-5.30} &16.86 &-3.92 \\
\hline
\multicolumn{2}{c}{\textit{Average}} &\textit{19.0} &\textit{-1.14} &\textit{18.7} &\textcolor{green}{\textit{0.35}} &\textit{19.4} &\textcolor{red}{\textit{-2.02}} &\textit{17.0} &\textit{-1.21} &\textit{18.5} &\textit{-0.31} \\
% & 19.00 & -1.54 & 18.71 &\textcolor{green}{-0.14} & 19.41 &\textcolor{red}{-2.03} & 17.00 &-1.38
\end{tabular}
}
\end{table}

Table \ref{tab:results} displays the average number of steps taken and units of energy collected and dropped for the two baseline agents (random walk and greedy search) and the three LLM-based agents (GPT-3.5-Turbo, GPT-4o, and GPT-o1-mini), evaluated using GRASP. Most \textit{Energy} results are negative numbers because they are averaged across all control settings, including a 0.3 unit energy cost per step that can lead to a negative energy result if the energy cost exceeds the energy collected. When considering the overall performance of each agent, we find that the greedy search agent performs the best across all settings, while GPT-3.5-Turbo performs the worst in 14 out of 14 settings. Additionally, GPT-4o performs better than GPT-3.5-Turbo, indicated by an unpaired t-test ($t(3198) = 6.78$, $p < 0.001$), but worse than GPT-o1-mini ($t(2398) = -6.27$, $p  < 0.001$) and is roughly on par with the random walk agent ($t(17598) = -0.48$, $p = 0.632$). All LLM-based agents perform worse than the greedy search agent ($p  < 0.001$), with only GPT-o1-mini performing better than the random walk method ($t(16798) = -7.64$, $p  < 0.001$). On the length of the steps, because the path of the random walk method is mechanically determined, it maintains consistent lengths of 19. The LLM-based agents exhibit consistent step counts within each individual model across different controls. However, when comparing their average length, GPT-4o consistently achieves the shortest path lengths, often between 16 to 17 steps, followed by GPT-0.1-mini, which has slightly longer paths, taking between 18 to 19 steps. In contrast, GPT-3.5-Turbo consistently produces the longest paths among the LLM-based agents, with about 19 to 20 steps.

Focusing on the performance within each control, we found that GPT-3.5-Turbo achieves the best \textit{Energy} results under the horizontally-skewed energy distribution, while GPT-4o performs the best when the energy distribution is random. Interestingly, GPT-o1-mini got the highest \textit{Energy} result under the spiral distribution. However, through paired t-tests, we found that the differences are not statistically significant ($p>0.01$). Additionally, we observed \textit{Energy} differences of 0.49 and 0.72 between the two settings of `Obstacle,' and `Energy Carrying Limit' for GPT-4o. This agent performs best without obstacles ($t(1598) = 3.03$, $p < 0.01$) and without any energy carrying limit ($t(1598) = 4.51$, $p  < 0.001$). For GPT-o1-mini, the difference between the two settings of `Energy Carrying Limit' is 1.11, suggesting that this agent performs best without any energy carrying limit ($t(798) = 4.62$, $p  < 0.001$). However, these differences are less apparent and not significant for GPT-3.5-Turbo, with the maximum difference being only 0.26 units of energy in the `Movement-related Action Set' control ($t(1598) = 1.45$, $p = 0.148$). 

All three LLM-based agents achieve notably lesser energy when given the constraint regarding the energy cost per step than when given no such constraint, with the minimum difference being 5.1 units of energy from GPT-4o ($p < 0.001$ for all LLM-based agents). Additionally, the `‘Movement-related Action Set' and `Starting Position' controls seem to play a lesser role in affecting the performance of GPT-3.5-Turbo, GPT-4o, and GPT-o1-mini, as the difference in the amount of energy collected and dropped is less apparent (with the maximum difference being less than 0.26 and $p > 0.1$ for all LLM-based agents).

% However, for GPT-3.5-Turbo, the `Energy Distribution' control shows a more significant impact, with a maximum difference of 0.68 between the `Spiral' and `Vertically-skewed' distributions. Turning to the average length of the steps taken by each agent, we observe that GPT-4o takes the least number of steps among the four agents, often between 16 to 17 steps, while GPT-3.5-Turbo takes the most number of steps, taking 19 to 20 steps. 

\begin{figure}[h]
\includegraphics[width=\textwidth]{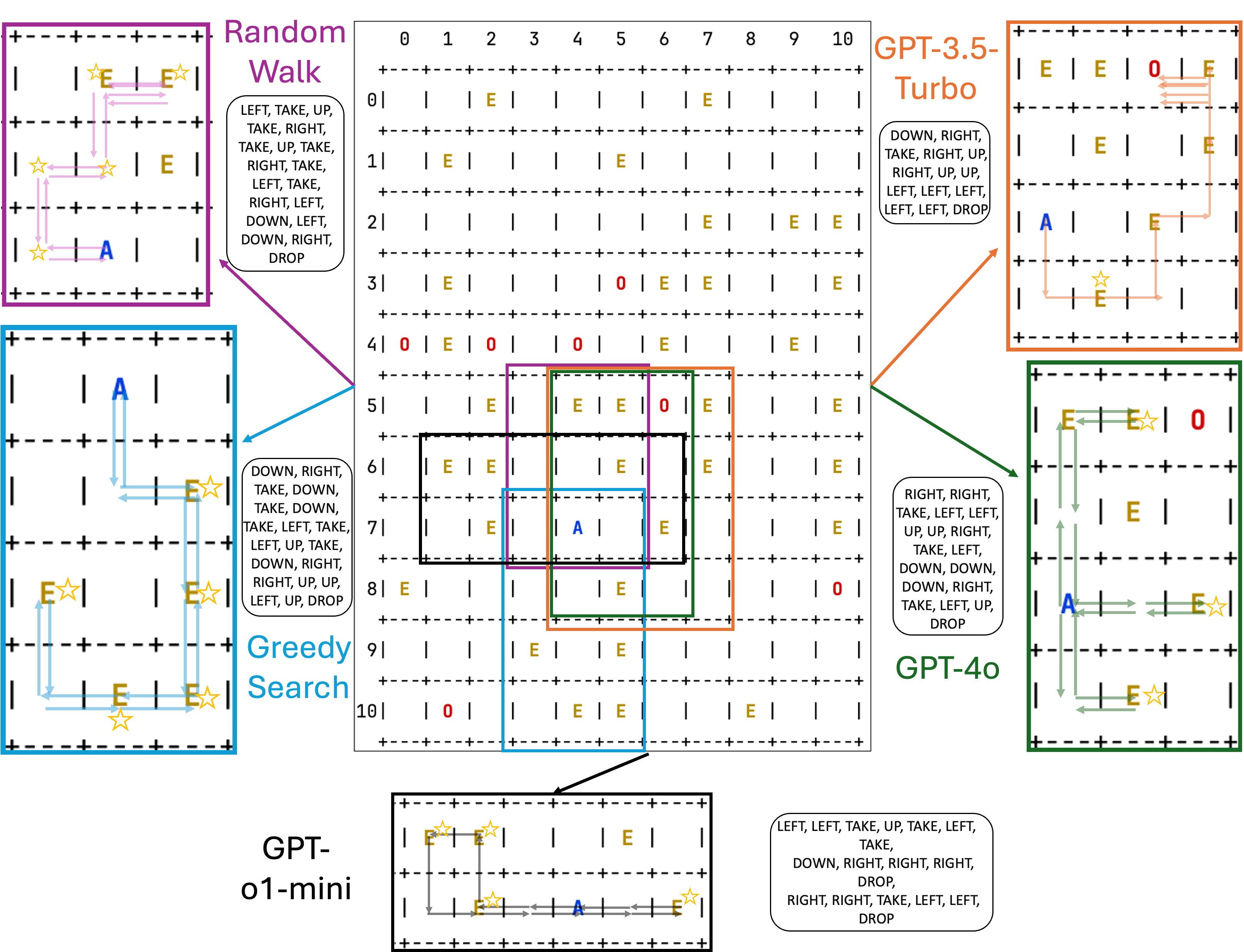}
\caption{
The path formed by the actions of the five agents (Random Walk, Greedy Search, GPT-3.5-Turbo, GPT-4o, and GPT-o1-mini) in the shown grid environment. The agents are under the following constraints: unlimited energy carrying limit, zero energy cost per step, and the $\mu_1$ movement-related action set (i.e., UP, DOWN, LEFT, RIGHT). Arrows indicate the movement-related actions, and a yellow star next to each cell signifies the agents' `TAKE’ action in that cell.
}\label{fig:result}
\end{figure}

Figure \ref{fig:result} illustrates a typical action path of the five agents within the shown instance of the grid environment. We observe that the random walk agent wanders through the grid without a clear objective and gives the `TAKE' action in cells without energy, while the greedy search agent exhibits a more object-oriented behavior, moving directly towards cells with energy and taking it. Focusing on the LLM-based agent, we find that GPT-3.5-Turbo includes cells with energy in its path but often fails to collect energy, repeatedly attempts to move through the cell with an obstacle, and does not return to the starting cell to drop the collected energy. In contrast, the GPT-4o agent demonstrates more intelligent behavior by actively going towards cells with energy and collecting them while avoiding obstacles, yet it still shows inefficient behaviors, such as missing nearby energy cells with energy on them and taking extra and unnecessary steps to achieve the same objectives. On the other hand, GPT-o1-mini exhibits even more sophisticated behavior in terms of both path efficiency and task completion. It prioritized collecting energy from cells in close proximity. However, we still observe a certain degree of inefficiency in that it sometimes overlooked nearby cells with energy or took longer paths than necessary to complete its objectives. These behaviors also reflect the performance of each agent observed in Table \ref{tab:results}.

\section*{Discussion}
In this paper, we introduce GRASP, a comprehensive benchmark designed to evaluate the Commonsense Spatial Reasoning (CSR) abilities of LLMs within structured grid environments. GRASP assesses agents’ abilities to navigate the grid environments and collect energy, presenting 16,000 diverse environments characterized by five types of energy distributions, two distinct obstacle configurations, two modes of agent starting position, and three kinds of agent constraints, including two sets of movement-related actions, two limits on the amount of energy that can be carried, and two different energy costs for each step. We evaluate the performance of two classic algorithms (random walk and greedy search) and three advanced LLMs (GPT-3.5-Turbo, GPT-4o, and GPT-o1-mini) using GRASP. Our findings show that GPT-3.5-Turbo under-performs compared to the random walk agent, often unable to successfully navigate the grid, avoid obstacles, and return to the starting point to drop the collected energy, suggesting a lack of CSR abilities. On the other hand, GPT-4o performs similarly to the random walk agent but is still less efficient than the greedy search agent. While GPT-4o can recognize spatial relationships and navigate around the grid environments, showing a primitive capability of CSR, its use of spatial information to come up with efficient plans remains limited. GPT-o1-mini, however, demonstrates superior performance compared to the random walk agent and other LLMs evaluated here. It is able to leverage spatial reasoning to plan efficient paths, actively prioritizing nearby energy cells and avoiding obstacles with a higher degree of accuracy. Despite its relative success, GPT-o1-mini still under performs when compared to the greedy search agent. For instance, it occasionally fails to optimize its route fully, leading to unnecessary or optimized steps.

% \hl{The following two paragraphs}

Our work builds upon and extends prior investigations into the spatial reasoning abilities of LLMs, moving beyond existing benchmarks to introduce a more practical evaluation framework. SpartQA \cite{mirzaee2021spartqa}, for example, evaluates spatial reasoning using a textual Question-Answering (QA) benchmark, which is constructed using grammar and reasoning rules to automatically generate a spatial description of visual scenes and corresponding QA pairs. Similarly, StepGame \cite{shi2022stepgame} is another multi-hop spatial reasoning benchmark that investigates the navigation ability of LLMs through complex spatial descriptions, while Liu et al. \cite{liu2022things} explored spatial commonsense by exploiting visual signals and evaluating LLMs' understanding of the relative scales of objects and the positional relationship between people and objects. On the other hand, Comsa et al. \cite{comșa2023benchmark} focused on reasoning with spatial prepositions, which is essential for accurately describing and interpreting spatial relationships. However, these benchmarks focus on interpreting and understanding spatial relationships in textual or visual formats, while GRASP evaluates the ability to utilize spatial information to accomplish related tasks. Our distinct approach assesses the practical application of spatial reasoning, requiring models to navigate and make plans based on spatial information. Another key distinction is that these benchmarks frequently depend on textual descriptions of scenes to convey spatial relationships. In contrast, our benchmark directly integrates text renderings of the environment to insert spatial information.

Additionally, platforms such as BabyAI \cite{chevalier2018babyai} provide a grid-based platform to support the research of including humans in the loop for grounded language learning. Additionally, Minigrid \cite{chevalier2024minigrid} provides modular and customizable grid-based environments designed for goal-oriented tasks in the reinforcement learning paradigm. However, these benchmarks primarily focus on evaluating an agent's ability to follow natural language instructions with spatial information, while our benchmark investigates the agent's ability to autonomously navigate the grid and collect resources using spatial information. More importantly, they see the task as needing further model optimization, while we argue that spatial reasoning should be treated as a commonsense ability and evaluated accordingly \cite{richardson2023commonsense}. Moreover, in our grid environment, the agent begins with complete knowledge of the grid’s layout and all necessary task information. Unlike other benchmarks that require agents to explore and actively gather data, our framework provides all information upfront, eliminating the need for sequential interaction with the environment.

Our study comes with several limitations. Firstly, the synthetic nature of GRASP’s grid environments may not fully capture the complexity of real-world CSR tasks, which include more dynamic and unpredictable environments with a greater variety of objects and incomplete or even hidden spatial information. Furthermore, GRASP focuses on commonsense spatial reasoning but does not consider the multi-modal nature of many real-world tasks where abundant visual information is available. Despite the current limitations and potential hallucination problems in LLMs’ multi-modal abilities, a reliable visual interpretation capability could provide the models with a richer set of spatial information than text alone. Also, we only consider scenarios with a single agent, whereas real-world environments often involve multiple agents, requiring more sophisticated behaviors, such as interacting and planning together with the other agents. Finally, our evaluation was limited to zero-shot prompting methods for LLMs, potentially underestimating their true capabilities.

To address these limitations, GRASP will be expanded in the future to include more dynamic environments that better simulate multi-modal CSR tasks, preferably developed using real-world data instead of purely synthetic ones. Additionally, with the rapid advancement of multi-modal foundational models, it is critical to integrate datasets that combine visual and textual spatial information to better capture the complexity of real-world CSR and provide a more comprehensive evaluation of the models. Extending the benchmark to scenarios involving multiple agents would also more accurately reflect the intricacies of real-world environments. Last but not least, investigating more sophisticated prompting methods could improve the performance of LLMs in these tasks, providing a more holistic evaluation of their commonsense spatial reasoning abilities.

In conclusion, GRASP is a novel benchmark for understanding and evaluating the commonsense spatial reasoning abilities of LLMs, albeit with some limitations. Our findings showcase the current limitations of even advanced LLM-based agents in efficiently completing CSR tasks and suggest the need for continued research and development in this area. As a structured and diverse benchmark, GRASP also provides a foundation for benchmarking future advancements in LLM capabilities, ultimately contributing to the development of more robust and intelligent systems capable of performing a wide range of commonsense planning tasks.

\section*{Methods}

\subsection*{Detailed Overview of Benchmark Construction}\label{sec:bench_con}

In general, the grid environment, $G$, can be understood as a 2D square array of cells $C_{ij}$, indexed from 0 to $k$ and from left to right and top to bottom, with $i$ representing row index and $j$ representing column index: {$C_{00}, ..., C_{0k}, C_{10}, ..., C_{1k}, ..., C_{k0}, ..., C_{kk}$}. A cell can either be empty ($C_{ij} = Null$), contain one unit of energy ($C_{ij} = E$), have an obstacle ($C_{ij} = O$), or be the starting point of the agent ($C_{ij} = A$). 

The agent is allowed pre-defined ($n$) steps. At each step, the agent can choose from a set of movement-related actions ($\mu$) that allows it to move around the grid environment and a set of resource-related actions ($\rho = \{TAKE, DROP\}$) that enables it to take the energy from or drop the collected energy to the cell that it is in. The agent's objective is to collect as much energy as possible in $n$ steps. 

\subsubsection*{Grid Environments}

We establish the side length $k$ of the grid environment to be 11. As a result, there are $11 \times 11$ = 121 cells in the grid. Based on this setup, we construct a set of grid templates that can be further instantiated to become a grid instance. Each grid template consists of three controls: Energy assignment, Obstacle construction, and Starting position. Each control can take several different values. 

Specifically, for the energy assignment, the \textit{random} distribution assigns an equal probability of having one unit of energy in each cell, setting a baseline test bed for the agent's ability. The \textit{vertically-skewed} and \textit{horizontally-skewed} distributions introduce asymmetry in the energy assignment, testing for any spatial bias in the agent's decisions. Furthermore, \textit{cluster} distribution introduces localized energy clusters, and \textit{spiral} distribution generates an energy path from the center, both exploring the agent's ability to identify and target specific regions or parts of the grid. Details about the five distinct types of energy assignment are described in the following:

% Obstacle construction adds another layer of complexity, which can stall the agent's movement, requiring more strategic navigation decisions. The choice of starting position, either inner or outer, further diversifies the grid setup, testing the agent's adaptability to different starting conditions.

% To comprehensively represent different possible energy distributions, we have developed five distinct types as follows: 

\begin{itemize}
    \item \textbf{Random} distribution of energy means that each cell in the grid has the same probability of containing one unit of energy. This probability is determined in during instantiation and follows a uniform distribution between 0.3 and 0.7. For one grid template, $P(C_{ij}=E)=p, p \sim U(0.3, 0.7), \forall i, j$.
    \item \textbf{Vertically-skewed} distribution means the cells in the top-half-part of the grid (rows 0 to 5) are more likely or less likely (determined in instantiation) to contain one unit of energy than the bottom-half-part of the grid (rows 6 to 10). In instantiation, the probability $p_{top}$ of a top-half-part cell containing one unit of energy is randomly instantiated from one of the two uniform distributions: $U(0.3, 0.4)$ and $U(0.6, 0.7)$. Then, the probability $p_{bottom}$ of a bottom-half-part cell containing one unit of energy is defined as $p_{bottom} = 1 - p_{top}$. Formally, $P(C_{ij}=E)=p_{top}$, $\forall$ $0<=i<=5, j$ and $P(C_{ij}=E)=p_{bottom}$, $\forall$ $6<=i<=10, j$. As for $p_{top}$, \[
p_{\text{top}} \sim \begin{cases}
U(0.3, 0.4) & \text{with probability } 0.5 \\
U(0.6, 0.7) & \text{with probability } 0.5
\end{cases}
\]
    \item \textbf{Horizontally-skewed} distribution is similar to the vertically-skewed distribution. According to this distribution, cells in the left-half-part of the grid (columns 0 to 5) are more likely or less likely (determined in instantiation) to contain one unit of energy than the right-half-part of the grid (columns 6 to 10). In instantiation, the probability $p_{left}$ of a left-half-part cell containing one unit of energy is randomly instantiated from one of the two uniform distributions: $U(0.3, 0.4)$ and $U(0.6, 0.7)$. Then, the probability $p_{right}$ of a right-half-part cell containing one unit of energy is defined as $p_{right} = 1 - p_{left}$. Formally, $P(C_{ij}=E)=p_{left}$, $\forall$ $i, 0<=j<=5$ and $P(C_{ij}=E)=p_{right}$, $\forall$ $i, 6<=j<=10$. As for $p_{left}$, \[
p_{\text{left}} \sim \begin{cases}
U(0.3, 0.4) & \text{with probability } 0.5 \\
U(0.6, 0.7) & \text{with probability } 0.5
\end{cases}
\]
    \item \textbf{Cluster} distribution creates a number of energy clusters around the grid. Each cluster enables the cells in a 3 by 3 grid to be filled with energy. The centers of the clusters are chosen randomly. Hence, it is possible for clusters to overlap each other. The number of clusters is randomly chosen between 3, 4, and 5 during instantiation. All other cells do not contain any energy. Formally, if the center of the cluster is denoted by the cell $C_{ab}$, then $P(C_{ij}=E)=1$, $\forall$ $a-1<=i<=a+1, b-1<=j<=b+1$ and $a \sim U\{0, 10\}$ and $b \sim U\{0, 10\}$.
    \item \textbf{Spiral} distribution creates a spiral-like energy trail starting from the center cell (the cell in row 5, column 5). The spiral progresses outward with variations in the angle and radius. The angle is adjusted with random variation, and the radius also varies with random fluctuations. The angle \(\theta\) at step \(i\) is given by: $\theta = \frac{i}{10} + \epsilon_\theta$. \(\epsilon_\theta\) is a random variation in the angle and $\epsilon_\theta\ \sim$ \(U(-0.2, 0.2)\). The radius \(r\) at step \(i\) is given by: $r = \frac{i}{11 \cdot 10 / (2\pi)} + \epsilon_r$, where \(\epsilon_r\) is a random variation in the radius and $\epsilon_r\ \sim$ \(U(-0.2, 0.2)\). The coordinates \((x, y)\) of each cell are then calculated as: $x = \text{int}\left( 5 + r \cdot \cos(\theta) \right)$, $y = \text{int}\left( 5 + r \cdot \sin(\theta) \right)$. The probability of a cell \((x, y)\) containing one unit of energy is 1 if it lies on the spiral path defined by the above functions and 0 if not.
\end{itemize}

In the instantiation process, each grid is assigned one specific energy distribution type. Each cell in this grid is constructed to have one unit of energy according to its associated probability that is described above.

In addition to energy, a grid template can also contain obstacles. If a grid is established to have obstacles, each cell of the grid has a 0.1 probability of being an obstacle. That is, $P(C_{ij}=O)=0.1, \forall i, j$. If a cell contains an obstacle, the agent is not allowed to cross the cell and take any energy from the cell. Finally, the agent's starting position is determined. There are two positions: inner position and outer position. The inner position is defined as any cell in the square formed by rows 3 and 7 and columns 3 and 7 (i.e., $C_{ij}, \forall$ $3<=i<=7, 3<=j<=7$). The outer position is defined as any cell in the grid that is not in the inner position (i.e., $C_{ij}, \forall$ $0<=i<=2$ or $8<=i<=10$, and $0<=j<=2$ or $8<=j<=10$). In instantiation, the agent's starting position configuration is randomly chosen between the inner position and outer position. Then, a specific cell in the chosen position configuration is selected uniformly at random as the starting position, ignoring any energy or obstacle that is already in that cell.

In total, we generated 100 instances for each combination of the five different energy distributions, two distinct obstacle configurations, and two modes of agent starting positions, resulting in $100 \times 5 \times 2 \times 2 = 2000$ grid instances.

\subsubsection*{Agent Constraints}\label{section:agent_cons}
Along with different grid environments, we also consider constraints that the agent needs to follow when accomplishing the task. First of all, we fix the number of steps the agent can take to be $n=20$ steps. Additionally, the agent cannot move beyond the boundary of the grid, and any invalid action will result in no change to the environment.

For each step, we design two sets of movement-related actions that the agent is allowed to take: $\mu_1=$ \{UP, DOWN, LEFT, RIGHT\}, $\mu_2=$ \{UP, DOWN, LEFT, RIGHT, UPLEFT, UPRIGHT, DOWNLEFT, DOWNRIGHT\}. The outcome of each action corresponds to the characteristics suggested by its name. For example, UP allows you to move one cell up in one step, and UPLEFT allows you to move diagonally one cell up and left in one step. The two sets of movement-related actions offer the agent varying degrees of movement flexibility, with one offering basic movements while the other allows for additional diagonal movements. Additionally, we design two resource-related actions: $\rho = \{TAKE, DROP\}$. TAKE allows the agent to take the energy from the cell if the agent is moved to a cell with energy, and DROP allows the agent to drop all the energy that the agent carries to the cell that the agent is currently on.

Furthermore, we design two constraints that could potentially affect the agent's behaviors. The first constraint, if active, adds a limit on the number of energy units that the agent can carry to a maximum of 2 units, requiring the agent to drop energy before acquiring any more of it beyond two units. The second constraint, if active, adds a cost to each action, forcing the agent to consider the energy expenditure of its actions. This cost is defined as 0.3 units of energy per action. Combining each of the 2000 grid environment instances with the three different agent constraints, we have a total of $2000 \times 2 \times 2 \times 2 = 16000$ different instances of the benchmark.

\subsubsection*{Grid Representation to An LLM-based Agent}\label{section:grid_repre}

So far, we have described the grid as an abstract environment. However, to present it to an LLM-based agent, the grid needs a concrete representation. Previous works either use text descriptions \cite{mirzaee2021spartqa, shi2022stepgame} or involve another modality (i.e., visual) \cite{chevalier2018babyai, chevalier2024minigrid}. While textual description can convey spatial information without visual elements, it comes with several disadvantages. For example, spatial relationships can be described in multiple ways, leading to bias in the way of description. Moreover, there is a limitation on the granularity of description, as it is unclear at what level should the spatial relationships be detailed. Describing complex environments can also result in excessive information, making it challenging to provide an accurate depiction. In contrast, visual representations offer detailed spatial information about the environment. However, they also present their own challenges, as the completion of tasks depends on the agent’s visual processing abilities, which can have inherent flaws. Moreover, visual representations can be overly complex, potentially confusing the agent

Our approach balances the benefits and drawbacks of both text description and visual representations by considering a hybrid representation. We present the grid as an 11 by 11 matrix labeled with coordinates from 0 to 10 on both the x-axis (top) and y-axis (left). The grid is constructed with cells separated by dashed lines, forming a series of rows and columns. The cells are separated vertically using ``+ - - - +" and horizontally by ``|". Each cell contains a character that represents if there is a unit of energy, an obstacle, or an agent on it. Specifically, a unit of energy is represented as ``E''; an obstacle is represented as ``O''; the agent is represented as ``A''. If the cell contains none of the three entities, it will be left as a whitespace character with length one. Additionally, to align the cells both horizontally and vertically, an extra whitespace character is added on both sides of the cell content. An example of the grid is shown in Figure \ref{fig:result}, and its text form is shown in Appendix A.1.

Our approach eliminates the need for textual descriptions by actively representing the spatial relationships within text space. It has been shown that LLMs are capable of understanding such representation \cite{patel2021mapping}. Furthermore, there is no additional visual processing capability requirement for LLMs, allowing the evaluation to concentrate on their ability to reason using explicit spatial information rather than their ability to interpret visual information. Our approach enhances the experimental control over biases and hallucinations.

\subsection*{Implementation Details for Experimental Study}

\subsubsection*{Baseline agents}
The random walk method randomly selects one out of the four or eight movement-related actions, depending on the available actions. Following each of the movement-related actions, this approach will select the TAKE action to take the energy from the cell, whether or not the cell has any energy. This `move and take' behavior will continue 6 times. Then, this approach will select the complement of the movement-related actions that were selected before in reverse order to get back to the starting position and finally select DROP to drop all the energy. The complement of a movement-related action is defined as the complementary action that exactly reverses the movement. For example, if the initial movement-related action was `UP,' the complementary action would be `DOWN.' In total, this will take 19 steps, which is within the required 20 steps.

The greedy search agent starts from the agent's current position and uses a Breadth-First Search algorithm to navigate the grid. It considers all four $\mu_1$ or eight $\mu_2$ possible movement-related actions, depending on the available actions. At each step, it checks the neighboring cells in all directions. The sequence of the neighboring cells in consideration is randomized. If a neighboring cell contains energy, the search stops, and the path to this cell is recorded. If all neighboring cells are either empty or obstacles, the search continues to the next set of neighboring cells. Once the nearest energy cell is found, the recorded path is used to move the agent step-by-step toward that cell. This will continue until a neighboring cell that has energy is too far from the current position, and the remaining number of steps is not enough for the agent to go there and get back to the starting position using the complement of past movement-related actions in reverse order. At this point, the agent simply uses the complement of past movement-related actions in reverse order to get back to the starting cell and drop all the collected energy. The pseudo-code of both methods is provided in the Supplementary Material.

\subsubsection*{Prompts used for evaluating the LLM-based agents}

The system prompt:
\begin{lstlisting}
You are an agent in a grid world. The grid world consists of cells. Each cell may have one unit of energy or no energy at all.{obstacle_prompt} The goal for you is to collect as much energy as possible and put the collected energy back in the cell where you started. You have 20 steps.{movement_prompt} You can collect energy from a cell by being in the cell and TAKE the energy from the cell. If there is no energy in the cell, you cannot take any energy from it.{energy_limit_prompt} You can not move across the boundary of the grid world. You can drop all your energy by DROP.{cost_of_step_prompt} You can use less than 20 steps. Any invalid step will not cause any change in the grid world.
\end{lstlisting}

The variables encapsulated in \{\} are depended upon the specific environment that the agent is in. If the grid has obstacles, then the obstacle\_prompt is ` Some cells are blocked by obstacles. You cannot move to or through these cells.' If the grid does not have obstacles, the obstacle\_prompt is left as blank. If the agent's movement-related action set is $\mu_1$, the movement\_prompt is ` For each step, you can choose UP, DOWN, LEFT, RIGHT, TAKE, and DROP. UP allows you to move one cell up in one step. The other movements are similar.' If the agent's movement-related action set is $\mu_2$, the movement\_prompt is ` For each step, you can choose UP, DOWN, LEFT, RIGHT, UPLEFT, UPRIGHT, DOWNLEFT, DOWNRIGHT, TAKE, and DROP. UPLEFT allows you to move diagonally one cell up and left in one step. The other movements are similar.' If the agent has no limit on the number of energy that it can carry, the energy\_limit\_prompt is left as blank. If the agent has a limit of 2 units of energy that it can carry, the energy\_limit\_prompt is ` You can only carry two unit of energy at a time.' If the agent has a zero cost per step, then the cost\_of\_step\_prompt is left as blank. If the agent has a cost of 0.3 unit of energy per step, then the cost\_of\_step\_prompt is ` Each step costs you 0.3 unit of energy.' 

The user prompt:

\begin{lstlisting}
You are given the following as the representation of the grid world, where A is you, E is energy{user_obstacle_prompt}:\n{GRID}Give your sequence of steps as a list. For example: [STEP, STEP, ...]
\end{lstlisting}

The variables encapsulated in {} are depended upon the specific environment that the agent is in. If the grid has obstacles, then the user\_obstacle\_prompt is `, O is an obstacle'. If the grid does not have obstacles, the user\_obstacle\_prompt is left as blank. The GRID variable is the text representation of the grid. 

If a grid environment has obstacles and the agent is allowed the movement-related actions set $\mu_1$ and is restricted to 2 units of energy carrying limit and has a 0.3 unit of energy cost per step, the agent in this grid environment is given the following prompt:

The system message:
\begin{lstlisting}
You are an agent in a grid world. The grid world consists of cells. Each cell may have one unit of energy or no energy at all. Some cells are blocked by obstacles. You cannot move to or through these cells. The goal for you is to collect as much energy as possible and put the collected energy back in the cell where you started. You have 20 steps. For each step, you can choose UP, DOWN, LEFT, RIGHT, TAKE, and DROP. UP allows you to move one cell up in one step. The other movements are similar. You can collect energy from a cell by being in the cell and TAKE the energy from the cell. If there is no energy in the cell, you cannot take any energy from it. You can only carry two unit of energy at a time. You can not move across the boundary of the grid world. You can drop all your energy by DROP.Each step costs you 0.3 unit of energy. You can use less than 20 steps. Any invalid step will not cause any change in the grid world.
\end{lstlisting}

The user message:
\begin{lstlisting}
You are given the following as the representation of the grid world, where A is you, E is energy, O is an obstacle:\n    0   1   2   3   4   5   6   7   8   9   10 \n  +---+---+---+---+---+---+---+---+---+---+---+\n 0|   |   |   |   | E |   |   | E |   |   | O |\n  +---+---+---+---+---+---+---+---+---+---+---+\n 1| E |   |   | E | E |   | E | E | E |   | E |\n  +---+---+---+---+---+---+---+---+---+---+---+\n 2| O | O | E |   | E | E | O |   |   |   | E |\n  +---+---+---+---+---+---+---+---+---+---+---+\n 3|   |   |   |   | O |   |   |   |   |   | E |\n  +---+---+---+---+---+---+---+---+---+---+---+\n 4|   |   | O | E | E |   |   |   | O |   |   |\n  +---+---+---+---+---+---+---+---+---+---+---+\n 5|   |   | E |   |   | O |   |   |   |   |   |\n  +---+---+---+---+---+---+---+---+---+---+---+\n 6|   | A |   | E |   |   |   | E | E |   |   |\n  +---+---+---+---+---+---+---+---+---+---+---+\n 7| E | E | E | E | O | E | E | E |   | E | E |\n  +---+---+---+---+---+---+---+---+---+---+---+\n 8| E | E | E | E |   | E | E |   | E | E | E |\n  +---+---+---+---+---+---+---+---+---+---+---+\n 9| E | E | E |   | E |   | E | O | E | O | O |\n  +---+---+---+---+---+---+---+---+---+---+---+\n10|   |   |   | E | O |   | E |   | E | E | E |\n  +---+---+---+---+---+---+---+---+---+---+---+\nGive your sequence of steps as a list. For example: [STEP, STEP, ...]
\end{lstlisting}

\section*{Data Availability}

All data generated or analyzed during this study are provided in a GitHub repository accessed at  (\url{https://github.com/jasontangzs0/GRASP}).

\bibliography{sample}

\section*{Supplementary Material}

\subsection*{Examples of grid instances}\label{app:grids}

\begin{figure}[h]
\centering
\includegraphics[width=\textwidth]{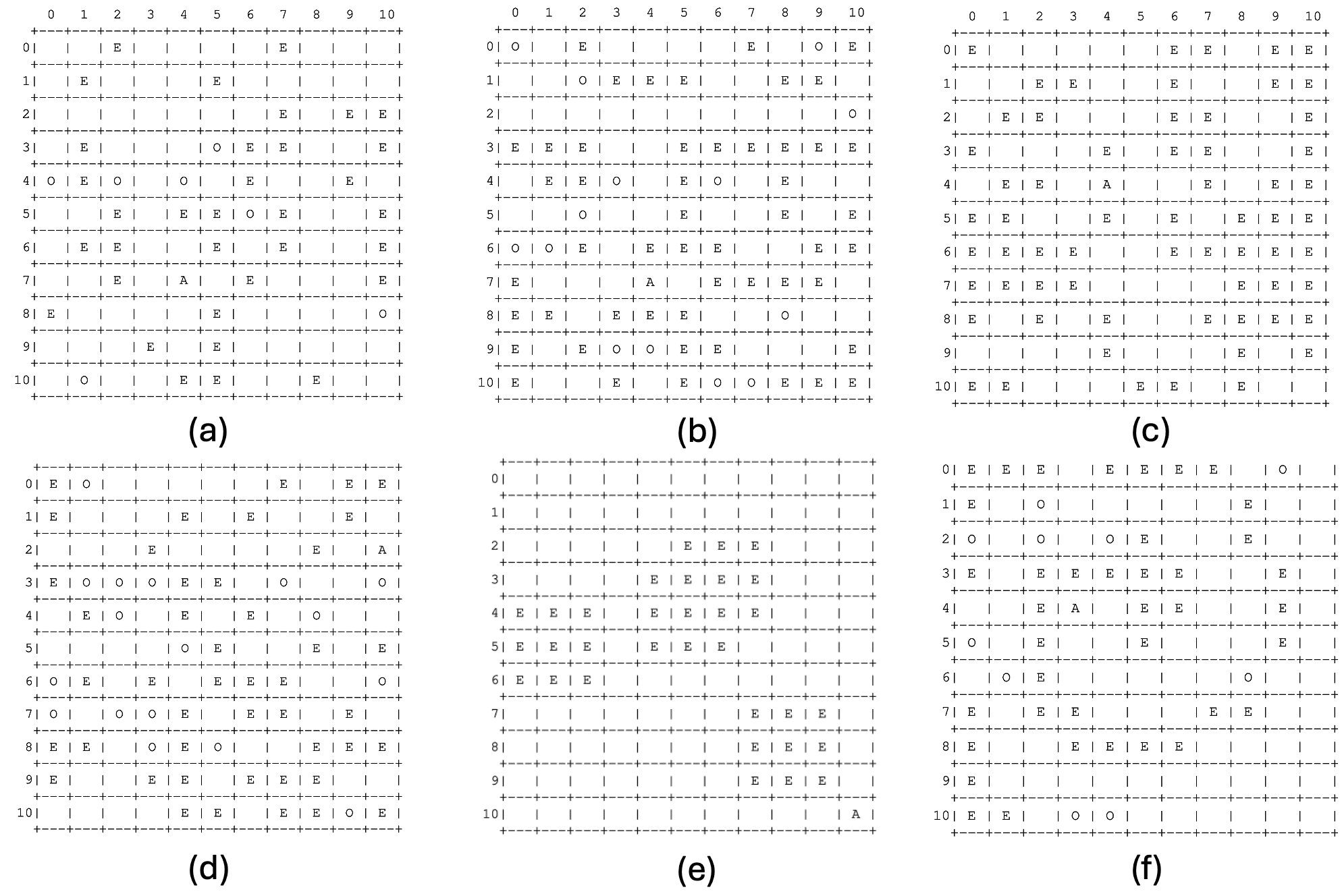}
\caption{
(a) The text form of the grid environment expressed in Figure \ref{fig:result}. The text form of the grid environment instantiated using: (b) random energy distribution with $p=0.52$, constructed with obstacles and using the inner position as the agent starting position; (c) vertically-skewed energy distribution with $p_{top}=0.38$, constructed without obstacles and using the inner position as the agent starting position; (d) horizontally-skewed energy distribution with $p_{left}=0.37$, constructed with obstacles and using the outer position as the agent starting position; (e) cluster energy distribution with 4 clusters, constructed without obstacles and using the outer position as the agent starting position; (f) spiral energy distribution, constructed with obstacles and using the inner position as the agent starting position.
}\label{fig:supp1}
\end{figure}

Examples of text forms and configurations of the grid environment shown in Figure \ref{fig:result} are reproduced in Figure \ref{fig:supp1}.

\subsection*{The pseudo-code for the two baseline agents}\label{app:pse}

Pseudocode of both baseline methods are reproduced under Algorithms 1 and 2 for completeness.

\begin{algorithm}
\caption{Random Walk Method}\label{algo:random}
\begin{algorithmic}[1]
    \REQUIRE The set of available movement-related actions $\mu$
    \STATE Initialize an empty list $actions$
    \FOR{$i \gets 1$ \textbf{to} $6$}
        \STATE Randomly select an action $a$ from $\mu$
        \STATE Append the selected action $a$ to $actions$
        \STATE Execute the selected action $a$
        \STATE Execute \textbf{TAKE} action
    \ENDFOR
    \FOR{$i \gets 6$ \textbf{to} $1$ \textbf{step} $-1$}
        \STATE Execute the complement of $actions[i]$
    \ENDFOR
    \STATE Execute \textbf{DROP} action
\end{algorithmic}
\end{algorithm}
\newpage

\begin{algorithm}
\caption{Greedy Search Method}\label{algo:greedy}
\begin{algorithmic}[1]
\REQUIRE The grid environment, agent's starting position, the set of movement-related actions $\mu$

\STATE Initialize an empty list $PastActions$
\WHILE{The remaining number of step > 0}
    \STATE Initialize queue $Q$ with the agent's starting position
    \STATE Initialize an empty set $Visited$
    \STATE Initialize an empty list $Path$
    \WHILE{$Q$ is not empty}
        \STATE Dequeue current position $current$ from $Q$
        \STATE Add $current$ to $Visited$
        \IF{$current$ contains energy}
            \STATE Record $Path$ to $current$
            \STATE Break
        \ENDIF
        \STATE Get a list of neighboring cells and their associated actions $(neighbor, action)s$ from $current$ that is one step away using the $action$ from the set of available actions $\mu$
        \STATE Randomize the order of $(neighbor, action)s$
        \FOR{each $(neighbor, action)$ in $(neighbor, action)s$}
            \IF{$neighbor$ is not in $Visited$ and $neighbor$ is not an obstacle}
                \STATE Enqueue $neighbor$ to $Q$
                \STATE Append $action$ to $Path$
            \ENDIF
        \ENDFOR
    \ENDWHILE
    \IF{energy cell found}
        \IF{remaining steps are not enough to get to the energy cell, return to the starting position, and drop the energy}
            \FOR{$i \gets length(PastActions)$ \textbf{to} $1$ \textbf{step} $-1$}
                \STATE Execute the complement of $PastActions[i]$
            \ENDFOR
            \STATE Execute \textbf{DROP} action
            \STATE Break
        \ELSE
            \FOR{$i \gets 1$ \textbf{to} $length(Path)$}
                \STATE Execute $Path[i]$
                \STATE Append $Path[i]$ to $PastActions$.
            \ENDFOR
            \STATE Execute \textbf{TAKE} action
        \ENDIF
        \STATE The remaining number of steps -= length($Path$) + 1
    \ELSE
        \FOR{$i \gets length(PastActions)$ \textbf{to} $1$ \textbf{step} $-1$}
                \STATE Execute the complement of $PastActions[i]$
        \ENDFOR
        \STATE Execute \textbf{DROP} action
    \ENDIF
\ENDWHILE
\end{algorithmic}
\end{algorithm}

% \subsection*{Complete data for the GRASP grid environments and experiments}

\subsection*{Selected negative results from GPT-4o-based agent}

\begin{figure}[h!]
\centering
\includegraphics[width=\textwidth]{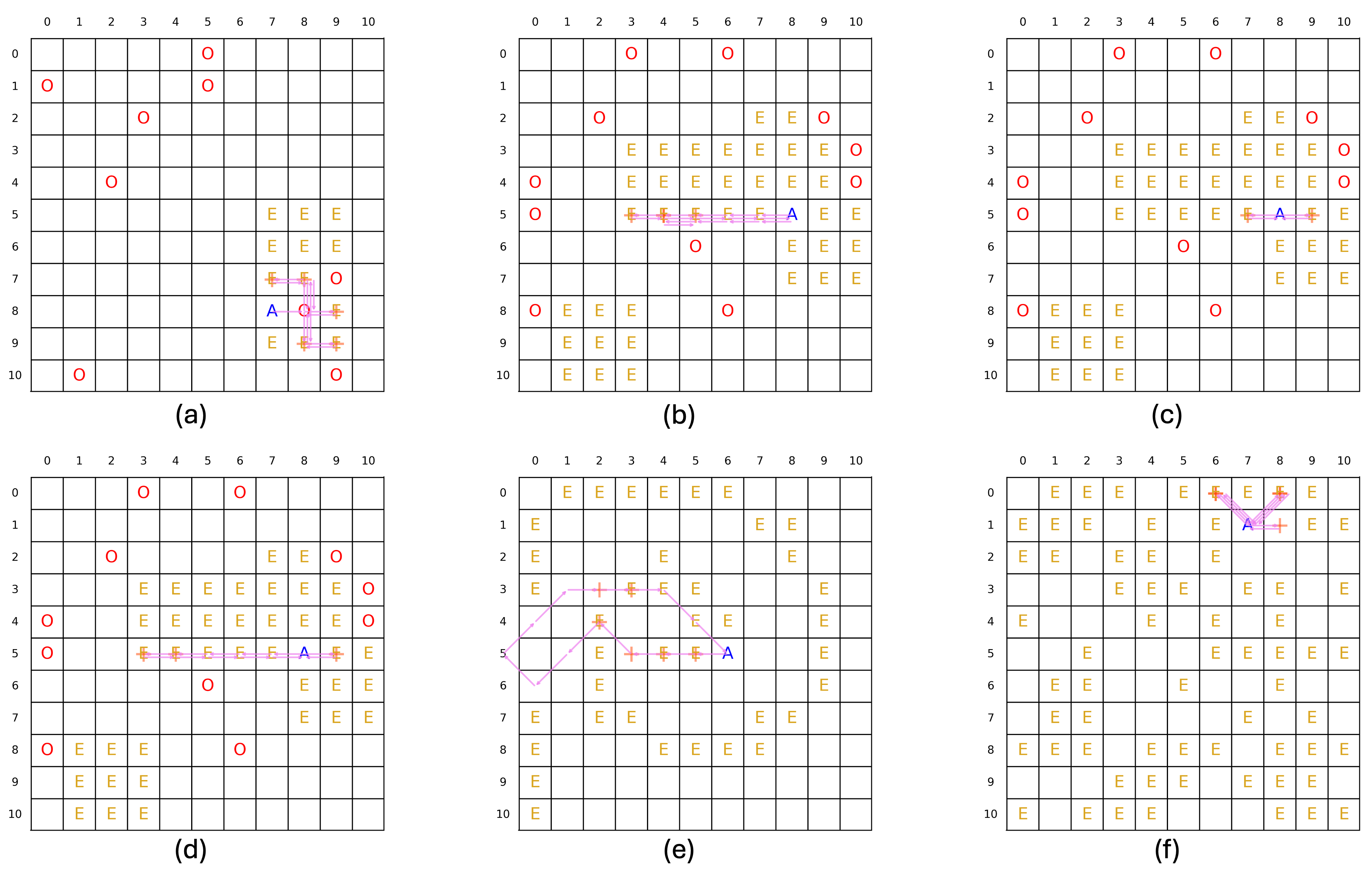}
\caption{
(a) The agent fails to avoid the obstacle. (b) The agent tries to take the energy from a distant cell instead of a nearby cell. (c) The agent is not utilizing all available steps. (d) The agent tries to take the energy from a distant cell instead of a nearby cell, even with an energy cost per step (e) The agent tries to go beyond the boundary of the grid. (f) The agent repeatedly goes back and forth between the same cells.}\label{fig:supp2}
\end{figure}

We provide six different instances where GPT-4o exhibits inefficient behaviors or even fails to obey the rule in Figure \ref{fig:supp2}.

\end{document}